\documentclass[twocolumn,showpacs,preprintnumbers,amsmath,amssymb]{revtex4}

\usepackage{graphicx}
\usepackage{dcolumn}
\usepackage{bm}

\begin{document}

\title{Problem Evolution: A new approach to problem solving systems}

\author{Goren \surname{Gordon}} \email{mail@gorengordon.com}
\author{Uri \surname{Einziger-Lowicz}}\email{Einziger.lowicz@gmail.com}


\begin{abstract}
In this paper we present a novel tool to evaluate problem solving
systems. Instead of using a system to solve a problem, we suggest
using the problem to evaluate the system. By finding a numerical
representation of a problem's complexity, one can implement
genetic algorithm to search for the most complex problem the given
system can solve. This allows a comparison between different
systems that solve the same set of problems. In this paper we
implement this approach on pattern recognition neural networks to
try and find the most complex pattern a given configuration can
solve. The complexity of the pattern is calculated using
linguistic complexity. The results demonstrate the power of the
problem evolution approach in ranking different neural network
configurations according to their pattern recognition abilities.
Future research and implementations of this technique are also
discussed.
\end{abstract}
\pacs{07.05.Mh}

\keywords{Genetic Algorithm, Neural Networks, Linguistic Complexity}
\maketitle

\section{Introduction}
\label{Introduction}
There is a variety of problem solving systems which offer
solutions to a given set of problems. Each system differs in the
mechanism in which it solves the problems within the problem
space. There is a need to rank the systems according to their
solving capabilities. A mere analytic way to compare the systems
is usually unavailable and a numerical approach usually uses
benchmark cases that fail to truly characterize the systems in a
more general way. The benchmarks are highly specific and sparse to
fully give an indication for the general problem and thus present
only a biased viewpoint of different solution ranking.

We present the problem evolution approach, which is a general,
unbiased and numerical method of evaluating problem solving
systems that allows a comparison between them. This approach ranks
the systems according to the most complex problem they can solve.
In order to implement the problem evolution method, a numerical
representation of the complexity of a problem within the problem
space must first be found. For each system, we use genetic
algorithm \cite{Goldberg} to tour the problem space in order to
find the most complex problem it can solve.

Any type of problem solving system has a different definition for
a complex problem according to the specification of the system.
Specifically, pattern recognition systems (e.g. neural network)
are required to successfully distinguish between as many different
patterns as possible. For this type of system, linguistic
complexity \cite{Trifonov} is a valid numerical characteristic for
the complexity of a problem, since it is based on the ratio
between the number of actual different elements used to the number
of maximal possible different elements. Thus, using genetic
algorithm to find the problem with the highest linguistic
complexity a given pattern recognition system can solve, can give
a good characterization for that specific system, and can enable a
comparison between different systems.

This paper begins with a brief description of genetic algorithms
and their special characteristics. It continues with a discussion
on multi-dimensional linguistic complexity and then describes
neural networks as a specific example of a pattern recognition
system, followed by the problem evolution process description and
results. The paper ends with a discussion and some philosophical
considerations.

\section{Genetic Algorithms}
\label{Genetic Algorithms}
The usage of genetic algorithms \cite{Goldberg} in optimization
problems is a rapidly developing field. Many applications of this
highly diverse method appear in software evolution \cite{Koza},
robotics evolution \cite{Cliff}, and evolvable hardware
\cite{Thompson}. In general, genetic algorithm uses the elements
of Darwinian evolution and natural selection. It is based on the
idea that a population of individuals exists and develops with
time. Each individual (genotype) codes for a specific function
(phenotype). Each generation, natural selection rules are
implemented on the population according to a fitness function,
which determines which individual functions best according to
predetermined constraints and desired objectives. Those that get a
high fitness score reproduce their genes into the next generation,
while others either copy themselves once or become totally
extinct. The fitness function can be described as a "landscape"
where high fitness is expressed as peaks in the landscape. The
individuals in the population tour the landscape and progress
upwards toward the local maxima using evolutionary operators,
namely, mutation and recombination \cite{Spears}. The former
meaning that random alterations in the genotype are performed,
whereas the latter addresses to sexual reproduction, where two
individuals exchange genetic information resulting in two
offsprings. Mutation causes "small movements" in the landscape,
whereas recombination creates "jumps" in order to avoid local
maxima in search for the global maxima of the fitness landscape.

Genetic algorithms are usually used to tour the solution space for
an optimized solution for a specific problem. The fitness function
for the individuals within the population is the proximity to the
optimal solution. Thus it facilitates an evolutionary process for
the best solution.

In our approach, each system has a different problem subspace
composed only of those problems it can solve. This subspace is not
known \textit{a-prior} and the genetic algorithm searches the
entire problem space. Each problem is presented to the system and
only if it can solve it, i.e. the problem is within the system's
subspace, the new genotype enters the population. The genetic
algorithm is used to find the most complex problem. Here, the
fitness function is the complexity of the problem. This is the
evolutionary process for the most difficult problem a given system
can solve.

\section{Linguistic Complexity}
\label{Linguistic Complexity}
Linguistic complexity \cite{Trifonov} is a simple and elegant way
of calculating complexity of strings of data. It is based on the
concept that the greater the vocabulary one uses, the more complex
the data. The complexity is the product of vocabulary usages of
all word sizes, where the latter measures the ratio of different
"words" used to the different "words" possible. The number of
possible words is limited by (1) the size of the alphabet and the
number of letters in the word (e.g. a 3-letter word with a binary
alphabet has $2^3=8$ possible different words) and (2) the length
of the string (e.g. a 3-letter word in a 6-letter string has
$6-3+1=4$ possible different words). For example, for the binary
string 010101, vocabulary usage of one-letter word is 2/2=1.0,
because two different letters appear (i.e. 0,1). The vocabulary
usage of two-letter words is 2/4=0.5, since only two different
two-letter words appear (i.e. 01, 10) whereas 4 combinations are
possible (i.e. 00, 01, 10 and 11). For a three-letter word, the
vocabulary usage is 2/4=0.5, since only two words appear (i.e. 010
and 101), but due to the short length of the string, only four
combinations are possible. Similarly, for four-letter words, the
vocabulary usage is 2/3 = 0.667. Thus, the complexity of the
string is $1.0\times 0.5\times 0.5\times 0.667=0.167$.

An extension to multiple dimensions was introduced lately and
enables to calculate linguistic complexity of any form of data
\cite{Gordon}. The complexity of the data set increases as more
different elements are presented.

As already noted, a numerical representation of a problem is
required to implement the problem evolution approach. The
multi-dimensional linguistic complexity is a general method for
acquiring it for many types of problems. Specifically,
multi-dimensional linguistic complexity is a good numerical
characteristic for pattern recognition. These systems are required
to distinguish between different patterns. A "problem" for pattern
recognition systems is a set of input-output pairs to be learned
and then distinguished. For example, a face recognition system has
to be able to match an image of a specific face (input) to a name
(output). By measuring the multi-dimensional linguistic complexity
of the images that system can detect, we ascertain the ability of
the system to distinguish between different elements. It gives us
a numerical value of the complexity of the problem that system can
solve.

Throughout the paper we will use a simple example, a
two-dimensional binary problem. The objective is to be able to
know for a specific coordinate in two dimensions (input) what
binary output to expect. Thus our problem is a binary image, and
its two-dimensional linguistic complexity is easily calculated
\cite{Gordon}.

\section{Neural Networks}
\label{Neural Networks}
Neural networks \cite{Haykin} (also referred in different
literature as artificial neural network, neurocomputers,
connectionist network, parallel distributed processors, etc.),
were introduced as an attempt to make an artificial computing
machine which simulate the human brain. Like the human brain,
neural networks are composed of many neurons, which are small
processing units that are capable of only simple calculations. The
power of this architecture is its parallelism of calculation and
its learning power.

A neuron receives input from other neurons, each having a
different weight representing the strength of the connection
between the given neurons. It sends the output of a simple
calculation of an activation function on those inputs to other
neurons. Neural networks have a standard structure which is
composed of layers of neurons, where each neuron in a layer
receives input from neurons in the previous layer and sends output
to the neurons in the next one. A given neural network has an
input layer, which receives input for the environment, an output
layer that produces the final result of the calculations and some
hidden layers.

There are many types of learning processes. We will discuss only
supervised learning \cite{Haykin}. The neural network can learn a
given data train set. This means that it can correlate between
inputs and outputs by changing the connection weights between
neurons in different layers. Later it is presented with a test set
and is marked in its ability to generate the correct output.

A neural network's configuration is described by its architecture
and the number of weights in it. The former is the arrangement of
the neurons in the different layers. The latter is the true number
of free parameters of the system. There is a simple correlation
between the two.

In practice, we will use neural networks to fully recognize a
binary image. The neural network will have two neurons in the
input layer (representing the input as the X, Y coordinate of a
pixel in the image), and one output neuron (representing the
binary "color" of that pixel). We will use different hidden layer
configurations throughout the paper. The training set will be
composed of all the pixels in the binary image, where the input is
the coordinate of the pixel and the output is its binary color. A
neural network is considered to fully recognize a given binary
image if it can learn the entire training set, i.e. if for all of
the image's pixels the neural network generates an output which
correlates to the binary color of that pixel.

Contrary to normal learning procedures, to determine whether the
neural network fails to fully recognize the image or slowly
converges to full recognition, a special stopping criterion is
required. We will use a convergence parameter which is the product
of the mean square error and the percentage of the pixels
accurately recognized. The learning process stops after a given
number of epochs, $N_c$, in which the convergence parameter fails
to decrease by a given fraction $\epsilon$.

\section{Problem Evolution for Neural Networks}
\label{Problem Evolution for Neural Networks}
We will now implement all the aforementioned techniques to
demonstrate the problem evolution approach, in order to evaluate
and compare different neural network configurations. We will
compare between configurations with similar architecture and then
between configurations with similar number of weights.
Two-dimensional linguistic complexity will be used as the
numerical representation of the problems. For each configuration,
genetic algorithm will tour the problem subspace the neural
network can solve to produce the most complex problem.

In order to find the most complex binary image a given neural
network can fully recognize we will use a genetic algorithm with
the following definitions. The population consists of binary
images the given neural network can fully recognize. The initial
population consists of simple binary images created by randomly
dividing the image to two homogeneous areas. Each new generation,
two new genotypes are created using recombination and mutation. If
a new genotype has higher linguistic complexity than the lowest
linguistic complexity of the population, the neural network can
fully recognize it and it does not appear already in the
population, it replaces the lowest genotype of the population. The
genetic algorithm ends after a given number of generations have
passed with no new genotype entering the population, i.e. when no
higher complexity genotype which the neural network can fully
recognize was found. Due to the possibility of reaching a local
maxima in the problem subspace instead of the global maximum,
several runs of the problem evolution are required.

Using this procedure gives us a tool for comparing different
configurations of neural networks. For each configuration the
problem evolution will result in a characteristic maximal
linguistic complexity. By comparing the latter one can determine
which configuration is better equipped to distinguish between a
greater number of elements within a binary image.

\section{Results}
\label{Results}
The problem evolution procedure used the following
parameters in all the detailed experiments:
\begin{itemize}
\item The genotypes are binary images consisting of 20x20 binary
pixels.

\item The genetic algorithm parameters are a population of 100
unique genotypes, a mutation rate of 0.0025 and a linear
recombination. The stopping criterion is a hundred generations of
no change in the population.

\item The neural network uses a $a\tanh(bx)$ activation function
with $a=1.7159$ and $b=2/3$ \cite{Haykin}; the $iRProp^+$ learning
process with commonly used parameters \cite{Igel}; and the
stopping criteria parameters were $N_c$=1000, $\epsilon$ =0.01.

\item In order to reduce the calculation time, each genotype
carries the weights of the neural network that can fully recognize
it. During the evolution operators (e.g. mutation and
recombination), the new offspring's weights were initialized using
the weights of its most similar parent. Thus reducing the number
of epochs needed to determine whether the offspring can be fully
recognized.

\end{itemize}

\begin{figure}[h]
\begin{center}
\includegraphics*[width=10cm]{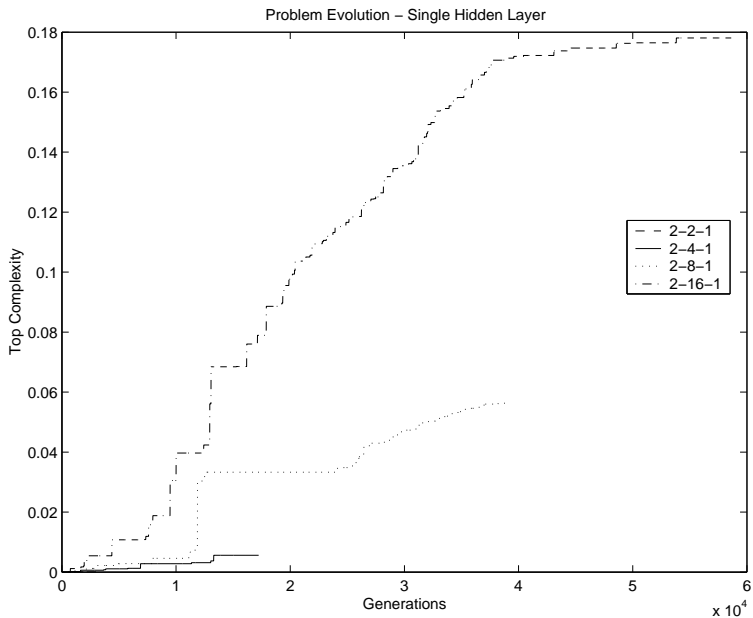}
\end{center}
\caption{\small \baselineskip=15pt Problem evolution of different
configurations of single layer neural networks. The graphs show
the complexity of the individual with the highest complexity in
the population in each generation.}
\label{SingleLayerFig}
\end{figure}

\begin{figure}[h]
\begin{center}
(a)
\includegraphics*[width=5cm]{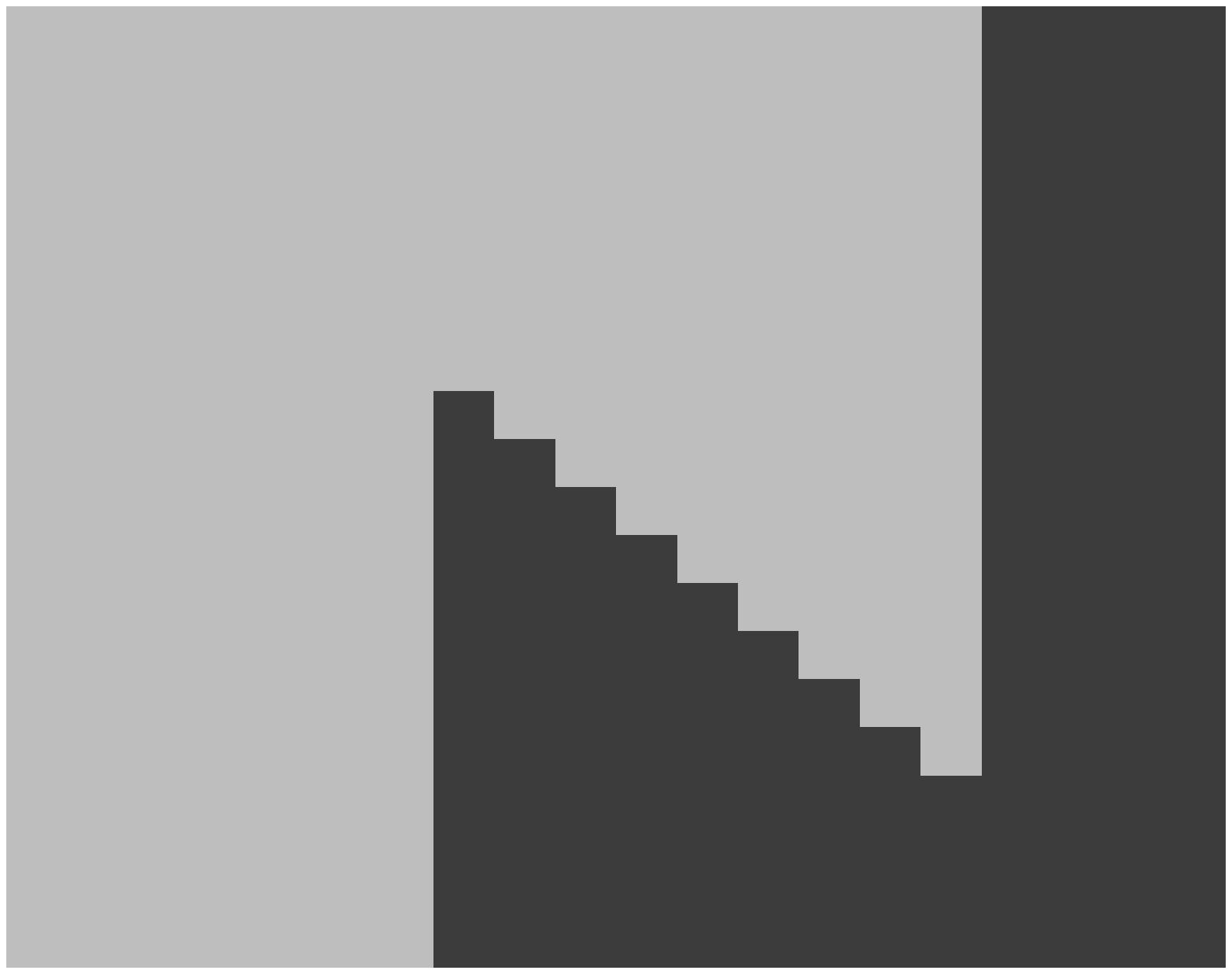}
\label{TopOrg2}
(b)
\includegraphics*[width=5cm]{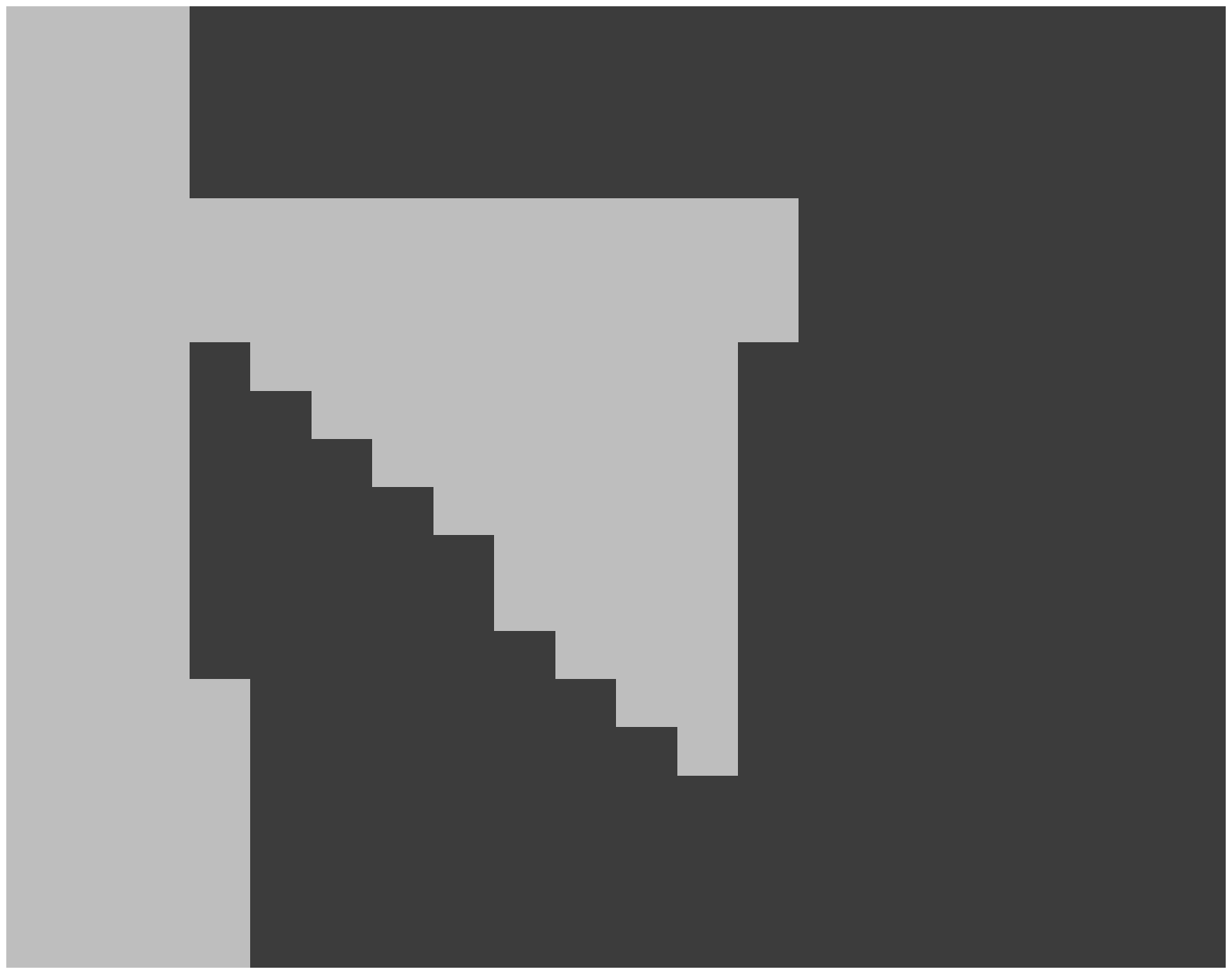}
\label{TopOrg4}
\\
(c)
\includegraphics*[width=5cm]{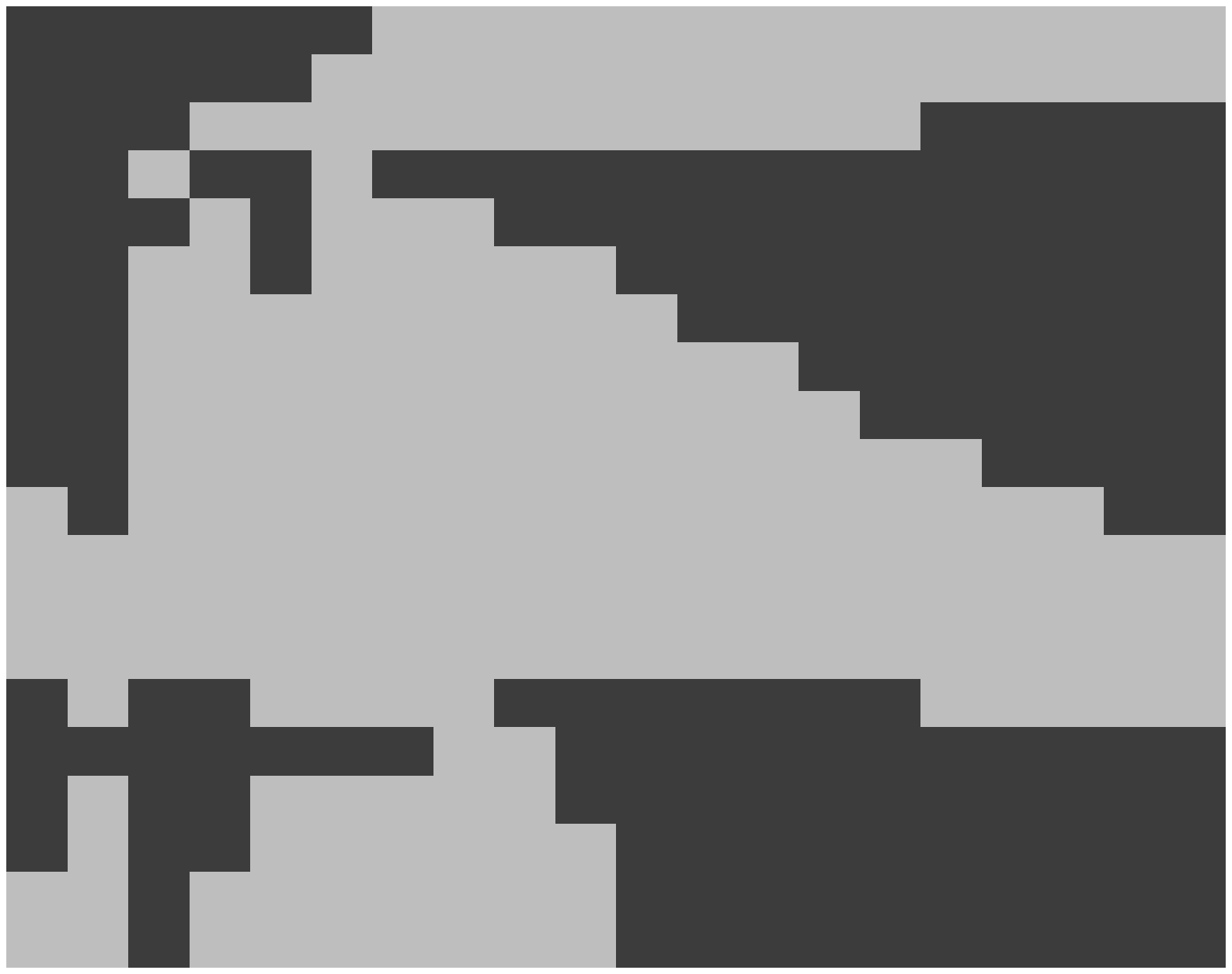}
\label{TopOrg8}
(d)
\includegraphics*[width=5cm]{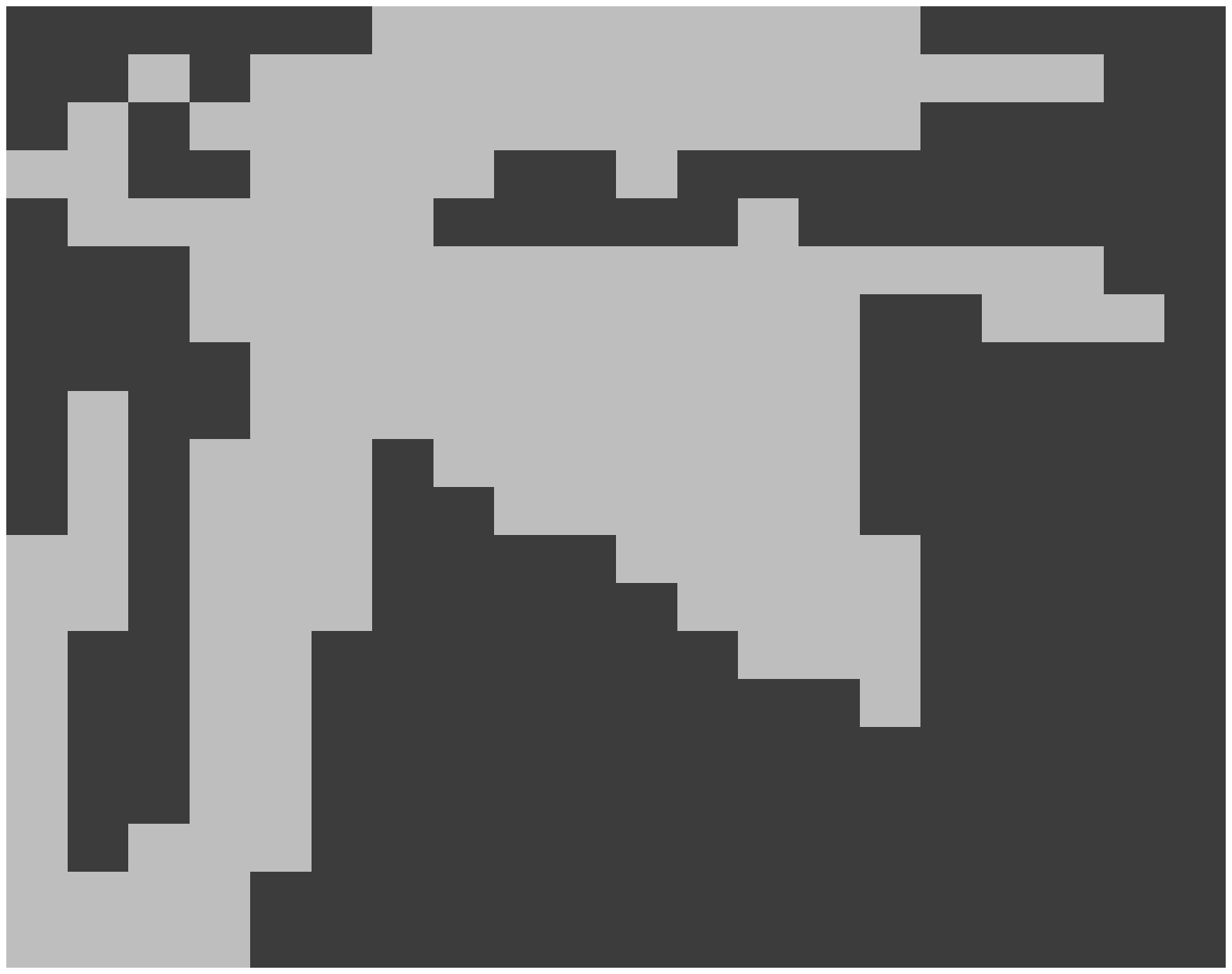}
\label{TopOrg16}
\end{center}
\caption{\small \baselineskip=15pt Highest complexity individuals
(20x20 binary images) the single hidden layer configurations could
fully recognize. (a) 2-2-1 configuration. Highest complexity:
0.000076. (b) 2-4-1 configuration. Highest complexity: 0.005611.
(c) 2-8-1 configuration. Highest complexity: 0.056301. (d) 2-16-1
configuration. Highest complexity: 0.178047.}
\label{TopOrgs}
\end{figure}

\textbf{\textit{Similar architecture.}} The aforementioned problem
evolution procedure was implemented on different configurations of
a neural network with a single hidden layer. The results are given
in Fig.~\ref{SingleLayerFig}, where the linguistic complexity of
the best genotype in each generation is plotted for different
numbers of neurons in the hidden layer. Each evolution process
ended after a hundred generations in which no new genotype
succeeded in entering the population (due to inability of the
neural network to fully recognize it or due to its too low
complexity), thus ending after a different number of total
generations. Fig.~\ref{TopOrgs} shows the individuals with the
highest complexity each configuration could fully recognize.

As can be seen, the maximal complexity reached is correlated to
the number of neurons in the single hidden layer. Thus
corroborating with the known fact that a neural network with a
higher number of neurons in its single hidden layer has a greater
pattern recognition capabilities.

\textbf{\textit{Similar number of weights.}} Next, we wish to
compare different configurations of neural networks with multiple
hidden layers, by comparing the maximal linguistic complexity
reached after the problem evolution process. In order to avoid
local maxima, the process was repeated three times and the maximum
of these runs was taken. Table~\ref{table1} shows a comparison
between different neural network configurations.

As can be seen, for the same number of weights, different
configurations have different capabilities in fully recognizing a
binary image. The number of neurons in the first hidden layer is
an important factor, as it determines the the output of the first
calculation. If it contains few neurons, information is lost.
Another important factor is the number of hidden layers (see
\cite{Chester}). A larger number of hidden layers enables
cascading of processes, i.e. calculation on the results of
previous calculations. With a given number of weights, there is a
tradeoff between these two factors. The results show that neither
a large single hidden layer configuration (2-8-1) nor three hidden
layers (2-3-3-2-1) gives the best recognition capabilities. The
best result is obtained by a two similar-sized hidden layers
(2-4-3-1).

\begin{table} %
\begin{tabular}{cc}
\hline
  Configuration (weights) & Maximal Linguistic Complexity \\
\hline
  2-2-6-1 (31) & 0.003 \\
  2-3-3-2-1 (31) & 0.042 \\
  2-8-1 (33) & 0.056 \\
  2-5-2-1 (30) & 0.135 \\
  2-4-3-1 (31) & 0.155 \\
\hline
\end{tabular}
\caption[]{Neural Networks' Maximal Complexity}\label{table1}
\end{table}

\section{Discussion}
The problem evolution process can be used in several ways. One can
find a correlation between the maximal linguistic complexity and a
neural network configuration. Thus for a given problem, one can
determine the best (i.e. minimal) configuration that solves that
problem by calculating the latter's linguistic complexity. This is
a novel approach for selecting the best neural network system for
a given problem. The advanced modern approaches for optimized
neural network learning are based upon growing and pruning neurons
in the process of learning \cite{Fahlman}. In our approach, one
can pre-determine the best configuration of neural network that
can learn the problem, thus minimizing the number of epochs needed
to be presented in-order to solve the given problem.

Another use of the problem evolution, as used in our research, is
to determine which neural network's configuration is better. By
finding the maximal linguistic complexity of each configuration,
one can compare different systems and determine which can solve a
more complex problem. In general, the problem evolution approach
can be used as an objective, numerical tool for comparing
different problem solving systems that solve the same problem. By
finding a numerical characteristic for the complexity of the
problem (e.g. multi-dimensional linguistic complexity), one can
use the genetic algorithm approach to evolve the most complex
problem a given system can solve, thus allowing the comparison of
different systems. This radical approach can be used as the final
tool in finding which system is the best in solving the problem
and which is obsolete.

On a more philosophical note, one can consider the brain as a
problem solving system and thus use the problem evolution process
to find the "most complex question" the brain can solve. By using
the genetic algorithm process to produce questions of increasing
complexity, one can converge to the most difficult question a
given person can answer.

\section{Conclusions}
This paper introduced a new perspective to evaluating problem
solving systems. It shifts the attention from the solution space
to the problem subspace and ranks the systems according to the
problems they can solve and not the solutions they produce.
Preliminary results were presented on evaluating neural networks
with different configurations. Further study is required in order
to obtain more precise guidelines of the optimal properties of
neural networks.

Problem evolution should be explored in new fields and implemented
on other problem solving systems in order to exploit its full
potential.

\end{document}